\documentclass[10pt,oneside,twocolumn,a4paper]{article}
\usepackage{hyperref}
\usepackage{graphicx}
\usepackage[utf8]{inputenc}
\usepackage{mathptmx}
\usepackage{tabularx}
\usepackage{ragged2e}
\usepackage[T1]{fontenc}
\RequirePackage[blocks]{authblk}
\usepackage{babel}
\usepackage{amsmath,amsfonts}
\usepackage{algorithmic}
\usepackage{algorithm}
\usepackage{array}
\usepackage{subfigure}
\usepackage{caption}
\usepackage{textcomp}
\usepackage{stfloats}
\usepackage{verbatim}
\usepackage{cite}

\newcolumntype{Y}{>{\centering\arraybackslash}X}
\usepackage{booktabs}
\usepackage{siunitx, arydshln}
\makeatletter
\let\ps@plain\ps@empty
\def\@xivpt{14bp}

\def\@sect#1#2#3#4#5#6[#7]#8{%
  \ifnum #2>\c@secnumdepth
    \let\@svsec\@empty
  \else
    \refstepcounter{#1}%
    \protected@edef\@svsec{%
      \ifnum #2<4
        \hb@xt@10mm{\csname the#1\endcsname}\relax
      \else
        \hb@xt@12mm{\csname the#1\endcsname}\relax
      \fi}%
  \fi
  \@tempskipa #5\relax
  \ifdim \@tempskipa>\z@
    \begingroup
      #6{%
        \@hangfrom{\hskip #3\relax\@svsec}%
          \interlinepenalty \@M #8\@@par}%
    \endgroup
    \csname #1mark\endcsname{#7}%
    \addcontentsline{toc}{#1}{%
      \ifnum #2>\c@secnumdepth \else
        \protect\numberline{\csname the#1\endcsname}%
      \fi
      #7}%
  \else
    \def\@svsechd{%
      #6{\hskip #3\relax
      \@svsec #8}%
      \csname #1mark\endcsname{#7}%
      \addcontentsline{toc}{#1}{%
        \ifnum #2>\c@secnumdepth \else
          \protect\numberline{\csname the#1\endcsname}%
        \fi
        #7}}%
  \fi
  \@xsect{#5}}
\renewcommand\LARGE{\@setfontsize\LARGE{16}{20}}
\def\abstract#1{\def\@abstract{#1}}
\def\abstractEn#1{\def\@abstractEn{#1}}
\def\titleEn#1{\def\@titleEn{#1}}
\def\@maketitle{%
  \newpage
  \null
  \let \footnote \thanks
    {\LARGE\bfseries\RaggedRight \@titleEn \par}%
    \vskip 1\baselineskip%
    {\normalsize
      \@author\par}%
    \vskip \baselineskip%
    {\section*{Abstract}
      \@abstractEn}%
  \par
  \vskip 3\baselineskip}

\renewcommand\section{\@startsection {section}{1}{\z@}%
                                   {-3.5ex \@plus -1ex \@minus -.2ex}%
                                   {\baselineskip}%
                                   {\normalfont\Large\bfseries\RaggedRight}}
\renewcommand\subsection{\@startsection{subsection}{2}{\z@}%
                                     {\baselineskip}%
                                     {1ex}%
                                     {\normalfont\large\bfseries\RaggedRight}}
\renewcommand\subsubsection{\@startsection{subsubsection}{3}{\z@}%
                                     {1\baselineskip}%
                                     {3bp}%
                                     {\normalfont\normalsize\bfseries\RaggedRight}}
\renewcommand\paragraph{\@startsection{paragraph}{4}{\z@}%
                                    {1\baselineskip\@plus1ex \@minus.2ex}%
                                    {3bp}%
                                    {\normalfont\normalsize\RaggedRight}}
\renewcommand\subparagraph{\@startsection{subparagraph}{5}{\parindent}%
                                       {3.25ex \@plus1ex \@minus .2ex}%
                                       {-1em}%
                                      {\normalfont\normalsize\bfseries\RaggedRight}}
\parindent\p@
\makeatother
\DeclareCaptionLabelSeparator{enskip}{\enskip}
\title{Beitragstitel (16 pt fett)}
\titleEn{Energy-efficient Deployment of Deep Learning Applications on Cortex-M based Microcontrollers using Deep Compression}
\author{Mark Deutel$^{1}$, Philipp Woller$^{2}$, Christopher Mutschler$^{2}$, and Jürgen Teich$^{1}$}
\affil{$^{1}$ Friedrich-Alexander-Universität Erlangen-Nürnberg (FAU), Erlangen, Germany \\$^{2}$ Fraunhofer IIS, Fraunhofer Institute for Integrated Circuits IIS, Nuremberg, Germany}

\abstractEn{Large Deep Neural Networks (DNNs) are the backbone of today's artificial intelligence due to their ability to make accurate predictions when trained on large data sets. With advancing technologies such as the Internet of Things, interpreting large amounts of data generated by sensors is becoming an increasingly important task. However, in many applications, not only the predictive performance but also the energy consumption of deep learning models is of great interest.
This paper investigates the efficient deployment of deep learning models on resource-constrained microcontroller architectures via network compression. We present a methodology for systematically exploring different DNN pruning, quantization, and deployment strategies for different microcontroller architectures, with a focus on low-power ARM Cortex-M-based systems. The exploration allows to analyze trade-offs between key metrics such as accuracy, memory consumption, execution time, and power consumption. We discuss experimental results on three different DNN architectures and show that we can compress them to less than 10\% of their original parameter count before their prediction quality degrades. This also allows us to deploy and evaluate them on Cortex-M based microcontrollers.}

\begin{document}

\maketitle

\section{Introduction}

Deep Neural Networks (DNNs) have become dominant in many applications that require autonomous decision making based on environmental information, including audio recognition \cite{hershey_cnn_2017}, image classification \cite{he_deep_2016, krizhevsky_learning_2009}, or human activity monitoring \cite{kautz_activity_2017}. DNNs are advantageous because they are easy to set up and can be trained to detect correlations even when faced with high-dimensional data.

However, the execution of DNNs is energy, resource, and time consuming \cite{sze_efficient_2017, canziani_analysis_2017}. In situations where the trade-off between resource constraints, execution time, and prediction quality is critical, DNNs often struggle to compete with classical machine learning approaches \cite{lane_squeezing_2017}. However, with trends such as \textit{smart devices} and the \textit{Internet of Things} (IoT), there is growing demand and interest in deploying DNNs on microcontrollers.

Deep compression is a relatively new area of research that deals with the compression of DNNs. Prominent techniques include DNN graph pruning \cite{lecun_optimal_1989} and weight quantization \cite{jacob_quantization_2017}. Their goal is to reduce the resource footprint of a DNN by reducing the number of trainable weights and computational complexity while preserving the original predictive performance. 

\begin{figure}[t!]%
    \centering%
    \includegraphics[width=.7\linewidth]{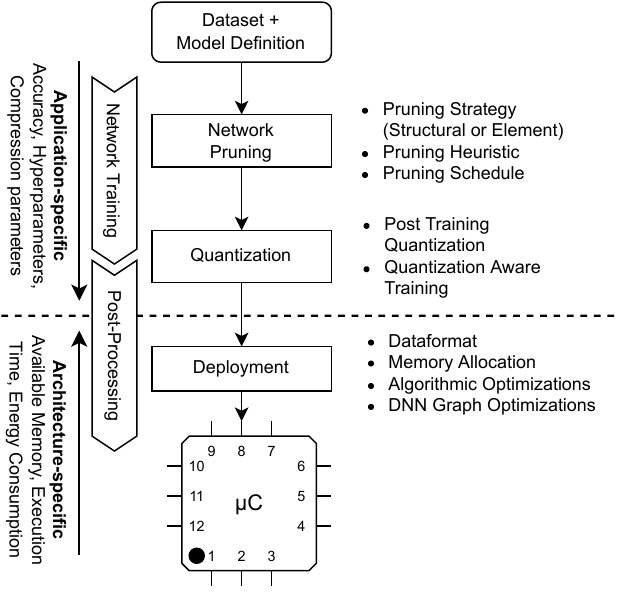}%
    \caption{Methodology overview.}
    \vspace{-2mm}
    \label{fig:pipeline_overview}%
    \vspace{-4mm}%
\end{figure}%

Based on these principles, several DNN compression pipelines have been proposed. Most notably, Han et al.~\cite{han_deep_2016} proposed a pipeline that combines network pruning, integer quantization, and Huffman coding. Others focus on quantization during network training~\cite{jacob_quantization_2017} or on structure-based pruning. This allows for immediate removal of pruned weights~\cite{li_pruning_2017, anwar_structured_2015}.
However, such well-established frameworks only trade off compression against prediction accuracy, but do not explicitly target energy efficiency and architecture-specific constraints such as memory availability and processing speed, which play an important role in many embedded applications.

This paper proposes a methodology for systematically training and deploying DNN architectures on Cortex-M-based microcontrollers. We present an automated pipeline that covers application-specific DNN training and compression, and combines it with target architecture-specific deployment, see Fig.~\ref{fig:pipeline_overview}. Our proposed pipeline consists of two main building blocks. First, from an application-specific perspective, we systematically explore compression techniques, i.e., network pruning and weight quantization, and configurations during the training of DNNs. Second, from an architecture-specific point of view, we realize the mapping from a high-level graph-based DNN representation to low-level code. This step involves an offline code generator and a runtime library. The former takes care of data layout, plans memory allocation, and outputs executable code, while the latter provides implementations of common DNN layers used by the generated code. Novel methods implemented in our proposed pipeline include up-front code generation and memory allocation planning, which eliminates the need for any form of network interpretation or dynamic memory management at runtime, and the use of sparse matrices generated as part of our pruning techniques using the \emph{Compressed Column Storage (CCS)} format \cite{barrett_templates_1994}.

In our experiments, we evaluate both compression and deployment for three common DNN architectures, namely AlexNet \cite{krizhevsky_imagenet_2017}, ResNet \cite{he_deep_2016}, and LeNet \cite{lecun_backpropagation_1989}. Our goal is to provide a thorough evaluation of the relationship between compressed DNNs and their predictive quality. In contrast to previous work, our results do not focus on deployment or compression alone, but provide detailed insight into the relationship between different pruning, quantization, and deployment strategies when applied in combination. We also deployed the compressed models on three target systems and discuss their memory consumption, execution time, and power consumption.

The rest of this paper is organized as follows. Sec.~\ref{sec:relatedwork} discusses related work. Secs.~\ref{section:pipeline} and~\ref{sec:arch_depl} provide details about our compression and deployment pipelines. Sec.~\ref{sec:eval} discusses our experimental results. Sec.~\ref{sec:conclusion} concludes.

\section{Related Work}
\label{sec:relatedwork}

Existing research primarily compresses DNNs through network pruning and weight quantization. These techniques are well understood, as research has been conducted to explore the effects of pruning and quantization on a network's predictive performance \cite{liu2018rethinking}. However, when applications are deployed for embedded targets, they are defined by the constraints imposed by the platforms they use. As a result, the suitability of DNN models for deployment on microcontrollers is determined not only by their accuracy, but also by their memory footprint and inference time. Therefore, this work extends existing findings by analyzing the effects of DNN compression, in particular DNN pruning and weight quantization, not only on accuracy, but also on relevant deployment metrics, i.e., memory consumption, latency, and power consumption.


Early work \cite{lecun_optimal_1989} introduced pruning as a measure to remove connections from a trained neural network using saliency scores based on second-order derivatives. This idea has been extended to perform pruning of CNNs at the filter or channel level \cite{anwar_structured_2015, li_pruning_2017} with the advantage of being able to remove pruned structures from DNNs. Furthermore, a major challenge in DNN pruning is to decide which structures to remove at what point during training. Existing research has focused on heuristics to computationally approximate the problem, focusing on metrics such as absolute size/l-norm \cite{han_learning_2015, li_pruning_2017}, gradient size \cite{molchanov_pruning_2017}, or statistics such as average percentage of zeros in activation tensors \cite{hu_network_2016}.

In addition to pruning, there have been efforts to further reduce the memory footprint and inference cost of DNNs in the form of weight quantization. Existing research has explored applying this technique either after DNN training is complete \cite{nagel2019data} or as part of the main training loop \cite{jacob_quantization_2017}. While the most commonly chosen target data type for quantizing DNNs is 8-bit integer, other commonly proposed data types are binary/low-bit integer types \cite{courbariaux_binaryconnect_2016} and fixed-point types \cite{lin_fixed_2016}.


Recent scientific work has provided some insight into the generalized use of DNNs on microcontrollers. The approach most closely related to our proposed methodology is MCUNet \cite{lin_mcunet_2020}. Similar to our pipeline, the authors describe a two-step process to seamlessly combine model design (TinyNAS) with an inference engine (TinyEngine). However, MCUNet differs from our approach in the way it generates suitable DNN candidates for deployment. To find networks that satisfy the constraints of the target platform, MCUNet focuses on neural architecture search (NAS) \cite{zoph_neural_2017}, while our framework starts from known existing DNN architectures and then dynamically scales them down during training using pruning and quantization techniques.

A more general approach to a microcontroller deployment framework is \emph{tfl-micro}~\cite{MLSYS2021_tflmicro}, which supports the execution of quantized tensorflow lite models on ARM Cortex-M-based hardware using ARM's \emph{CMSIS} library. However, this also limits the framework as it requires the use of \emph{tensorflow (TF)} for model training and only supports a subset of the features implemented in TF.



\section{Compression and Deployment Pipeline}
\label{section:pipeline}

Our pipeline is fully integrated and seamlessly covers the complete DNN training and deployment process. Our methodology uses both network pruning (Sec.~\ref{subsection:pruning}) and weight quantization (Sec.~\ref{subsection:quantization}), both of which can be controlled by a set of additional \emph{hyperparameters}. Furthermore, the trained and compressed DNNs are directly converted from their graph-based representation to executable architecture-specific program code (see Sec.~\ref{sec:arch_depl}). As a result, our pipeline can easily be integrated with existing meta-heuristic optimization frameworks (e.g. Optuna \cite{akiba_optuna_2019}) to conduct design space exploration.

\subsection{Network Pruning}
\label{subsection:pruning}

The pipeline implements configurable elements for network pruning, i.e., (1) pruning techniques, (2) pruning heuristics, and (3) pruning schedule, which we describe in the following.

\textbf{Pruning techniques.} Pruning DNNs by removing parameters has been proposed early~\cite{hanson_comparing_1988, lecun_optimal_1989}. While initially being introduced to improve generalization and convergence, it has recently become a standard size reduction technique with little to no cost in accuracy. Our pipeline implements \emph{element-wise pruning} and \emph{structural pruning}. Element-wise pruning removes connections from a DNN's compute graph, i.e., parameters of the network are set to zero. Thus, these parameters do no longer influence the training error and are removed from the scope of the optimizer that trains the network. Structural pruning sets whole structures of parameters to zero. This has proven to be very efficient for pruning filters~\cite{li_pruning_2017} or channels~\cite{anwar_structured_2015} of 2D-convolutional layers, but it can also be applied analogously to rows and columns of linear layers. Its main advantage is that it removes entire structures from the weight tensors at once, resulting in a significant immediate reduction in parameters (in contrast to element-wise pruning, which produces sparse weight tensors).

\textbf{Pruning heuristics.} A critical aspect of pruning is selecting the elements or structures that, when removed, will have the least impact on predictive performance. 
A number of heuristics have been proposed that approximate optimal element-wise or structural pruning. In our framework, we have implemented many popular approaches that are based on different criteria such as magnitude \cite{han_learning_2015}, L-norm \cite{li_pruning_2017}, gradient \cite{molchanov_pruning_2017} or percentage of zeros found in activations \cite{hu_network_2016} to rank parameters or parameter structures by their approximated importance.

\textbf{Pruning schedules.} The pruning schedule determines when, how often, and to what extent a DNN will be pruned during training. We implement two well-known approaches: \emph{One-Shot Pruning}~\cite{lecun_optimal_1989} and \emph{Iterative Pruning}. One-shot pruning first trains a DNN until it achieves a \textit{reasonable} accuracy on the test dataset, and then prunes the DNN (optionally followed by a few epochs of re-training). Iterative pruning prunes a DNN over the course of training, allowing for an interleaved re-training. Thus, not all weights are removed at once, but gradually over several pruning iterations (eventually enforcing maximum sparsity). We implemented \emph{Automated Gradual Pruning (AGP)}~\cite{zhu_prune_2017}, which gradually increases the number of pruned weights $s_t$ starting at $t_0$ from an initial sparsity $s_i$ to a final sparsity $s_f$ over $n$ steps:
\begin{equation}
s_t=s_f+(s_i-s_f)\left(1-\frac{t-t_0}{n\Delta t}\right)^3,\ t \in \left\{t_0, \dots, t_0+n\Delta t\right\}.
\label{eq:agp}
\end{equation}

\subsection{Weight Quantization}
\label{subsection:quantization}

Quantization reduces the numerical resolution of the parameters and their computation. This not only reduces the memory footprint but also the computational complexity of a DNN. However, since parameter quantization introduces an additional error into the predictive performance, a major challenge is to fing a good trade-off between predictive quality and parameter resolution.

Our framework uses an \emph{affine mapping} scheme that transforms an original floating-point parameter into an 8-bit unsigned integer\footnote{See also~\cite{developers_onnx_2021} and \url{https://onnxruntime.ai/docs/how-to/quantization.html}.}. 
We apply a function $f(x)$ in combination with additional sets of trainable scale and zero point parameters:
\begin{equation}
\begin{split}
f(x) &= g\left(\left\lfloor\frac{x}{s}\right\rceil+zp\right),\\ s &= \frac{max_{data} - min_{data}}{255},\ 0\leq zp\leq 255,
\end{split}
\label{eq:quant_scheme}
\end{equation}
where $g(x)$ is the clamp-function to avoid data type overflows:
\begin{equation}
g(x) = \left\{ \begin{array}{rcl}x & \mbox{if} & 0\leq x\leq 255 \\ 255 & \mbox{if} & x>255 \\ 0 & \mbox{if} & x<0. \end{array}\right.
\label{eq:quant_scheme_clamp}
\end{equation}
The scale parameter $s$ defines the step size of the quantization, which is calculated as the ratio between the range over which values are distributed in the original floating-point space and the range covered by the quantized integer space. The zero-point parameter $zp$ denotes an exact point in the quantized space that represents zero in the original floating-point space. The two parameters can be defined either per tensor or per structure.

Quantization can be applied not only to weight tensors but also to activation tensors. We call this as \emph{full integer quantization}. During execution, most computations can then be performed in integer instead of floating-point space, which is beneficial for target systems that are not equipped with floating-point units. We give an example of applying full-integer quantization to matrix-multiplications. The general form is defined as:
\begin{equation}
\begin{split}
c_{ij} = \sum_{k=0}^n a_{ik} \cdot b_{ki}, \forall i \in \{0,\dots,m\}, \forall j \in \{0,\dots,p\},
\end{split}
\label{eq:matmul_quant}
\end{equation}
where the first line describes how the elements of a matrix $C$ are calculated from the elements of a $m \times n$ matrix $A$ and a $n \times p$ matrix $B$. In a fully-quantized DNN, both matrices $A$ and $B$ contain integer values and we must first de-quantize them by rearranging Equation~\ref{eq:quant_scheme} before multiplying them. Since the resulting matrix $C$ is in un-quantized space, we must quantize it by applying Eq.~\ref{eq:quant_scheme} again. By substituting and rearranging the previous computations we obtain
\begin{equation}
c_{ij} = g\left(zp_c +\left \lfloor{ \left(\frac{s_a \cdot s_b}{s_c}\right) \sum_{k=0}^n (a_{ik} - zp_a) (b_{ik} - zp_b) }\right \rceil\right).
\end{equation}
Note that only the scale parameters $\{s_a, s_b, s_c\} \in \mathbb{R}$ while all other parameters $\in \mathbb{N}_0$.

Our pipeline implements two common ways to determine when to apply quantization to a DNN. The first method quantizes as a post-process (PPQ)~\cite{hubara_improving_2020}, i.e., after training, and the second method integrates quantization into the training loop. The latter is called \emph{Quantization-aware Training (QAT)}~\cite{jacob_quantization_2017}. Both techniques have their advantages and disadvantages: PPQ is extremely easy to integrate, as it can be performed completely decoupled from a DNN's training process and does not require any invasive changes to a DNN's architecture (i.e., no retraining to fine-tune quantization parameters). However, this usually comes at the cost of a larger error introduced by the quantization, since the required scale and zero point parameters are only roughly approximated. In contrast, QAT adjusts the quantization parameters as part of the DNN's training process and can therefore produce better results. However, QAT only works properly with extensive network augmentation, which results in a more complex and computationally expensive training process.

\section{Architecture-Specific Deployment}
\label{sec:arch_depl}

\begin{figure}[t!]%
    \centering%
    \includegraphics[width=\linewidth]{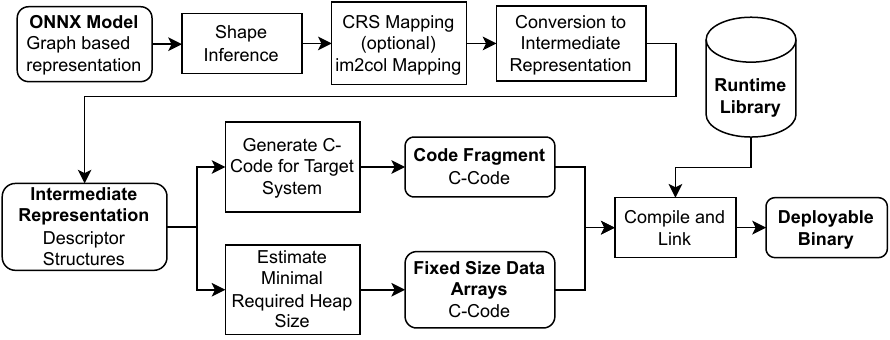}%
    \caption{Overview of the process implemented by our conversion tool, mapping trained ONNX DNN models to C-Code and finally compiling them to deployable binaries.}%
    \vspace{-2mm}%
    \label{fig:deploy_overview}%
\end{figure}

Our pipeline provides a deployment framework for targeting microcontrollers, see Fig.~\ref{fig:deploy_overview}. We call this framework \emph{dnnruntime}. It uses a platform-independent, offline, and ahead of time conversion tool together with a runtime library. The conversion tool maps pre-trained DNNs stored in the ONNX format to C code (Sec.~\ref{sec:conversion tool}), while the runtime library implements platform-specific DNN operators that are subsequently used by the code emitted from the conversion tool (Sec.~\ref{sec:runtime library}).
Our implementation is novel in that it exploits the static properties of trained DNNs (i.e., fixed layer configurations and parameters), thus eliminating the need to interpret the DNN at runtime. This includes dynamic memory allocation for intermediate tensors, which can be simulated offline, allowing heap allocation to be done at compile time. This not only reduces the computational overhead at runtime, but also allows metrics such as simulated memory consumption to be fed directly back into the overall optimization process without having to evaluate the model on the target system.

\subsection{Conversion Tool}
\label{sec:conversion tool}

The main functionality of the conversion tool is to generate ANSI C code based on a given ONNX model of the DNN to be deployed. This involves two steps: (1) parsing and converting the model to an intermediate representation, and (2) using this representation to determine a suitable data format, simulate memory allocation and generate an implementation describing the model's structure in code.

The ONNX format stores the computational graph of a DNN as a directed, acyclic graph with a set of fixed inputs and outputs. Each node in the graph represents an operation and can have multiple incoming and outgoing edges associated with it. The edges describe the flow of data during the execution of the DNN. Additionally, based on the type of operation a node represents, additional static parameter tensors can be assigned to it.

\subsubsection{Mapping ONNX to intermediate format}

First, we map a given ONNX representation to an architecture-specific intermediate format that can be used later to output program code. This involves three successive steps. 

First, we concatenate the static tensors of all nodes into a byte stream. The individual elements of each tensor are stored in the stream using little-endian byte order, as this is the default storage format on ARM architectures (of course, this can easily be changed). Additionally, we add padding bytes where necessary to avoid triggering the memory protection unit (MPU) when accessing tensor data at runtime. We then generate descriptor structures containing the location of each tensor in the byte stream and additional metadata such as data types and tensor shapes. Sparse tensors are treated as edge cases because they are generated by element pruning during the compression stage of our pipeline. To reduce memory consumption, our tool applies a conversion from the original full-size layout of the tensors to a more compact \emph{Compressed Row Storage (CRS)}~\cite{barrett_templates_1994} layout, see Fig.~\ref{fig:crs_mapping} for an example. CRS reduces the memory footprint, allows an optimized implementation of matrix-vector products, and does not impose any requirements on how sparsity is distributed within a tensor (hence, element pruning can ignore the subsequent space-saving storage of pruned tensors).

\begin{figure}[t!]
	\centering
	\scriptsize
	\subfigure{$\displaystyle A = \begin{bmatrix}
		10 & 0 & 0 & 0 & 1 \\
		0 & 7 & 0 & 2 & 0 \\
		0 & 0 & 8 & 0 & 0 \\
		14 & 0 & 0 & 0 & 6
	\end{bmatrix}$}
	\hfill
	\subfigure{\begin{tabular}{l c c c c c c c}
		\toprule
		\textbf{values} & 10 & 1 & 7 & 2 & 8 & 14 & 6 \\
		\midrule
		\textbf{col\_ind} & 0 & 4 & 1 & 3 & 2 & 0 & 4 \\
		\textbf{row\_ptr} & 0 & 2 & 4 & 5 & 7 & & \\
		\bottomrule
	\end{tabular}}
	\caption{Conversion of an asymmetric matrix $A$ (above) into its CRS representation (below): we emit three smaller arrays to the bytes-stream instead of one. The first one contains all non-zero values, the second one contains column indices, and the third one contains row pointers.}
	\label{fig:crs_mapping}
	\vspace{-4mm}%
\end{figure}

Second, the conversion tool generates descriptor structures for all dynamic activation tensors. This is more complicated than for static parameter tensors because activation tensors are represented in the ONNX model's compute graph only as edges. Edges are not required to provide any meta-information such as data types or shapes. However, this information is mandatory for our conversion tool. Therefore, we implement a process called \emph{shape inference}. The idea is to trace the execution of a DNN through its compute graph from input to output nodes, and use these traces to infer the shapes and types of intermediate tensors assigned to the inner edges.

Third, we parse and interpret all operator nodes in the ONNX computational graph and place them in a topological and serialized order. This information is used during code generation to determine the execution order of operations.

$~~$
\subsubsection{Code Generation}

Using the intermediate representation generated in the first step, the conversion tool can output code. We start by estimating the minimum heap size needed to store activation tensors. This information can be queried offline, since once a DNN is trained, its structure and dimensionality remain unchanged throughout its lifetime. Using the minimum heap size, we define a fixed size memory balloon at compile time (eliminating the need for dynamic memory management at runtime). A naive approach to estimating the size of this balloon computes the product of the shapes of all activation tensors and multiplies them by the byte sizes of their respective data types. However, this is not space efficient, since the lifetimes of these tensors are usually quite short. Therefore, parts of the heap memory can be reused for multiple tensors during an inference pass. This can have a large impact on the amount of memory required.

We take advantage of this by implementing an offline \emph{memory planning algorithm} based on graph tracing and using a first-fit allocation strategy therein. We estimate optimal heap re-usage in two steps:
First, based on the incoming and outgoing edges of nodes (i.e., operators) in the input model, the algorithm generates two lists per operator: The first list contains all tensors to be allocated for that operator (i.e., the allocation list), and the second list contains all previously allocated tensors that can be discarded (i.e., the discard list). Second, the algorithm iterates through the sequence of operators, starting with an empty, infinitely large memory balloon. For each operator, it first iterates the tensors in the corresponding free list and marks their space in the balloon as free. Then it iterates the allocate list and tries to reserve pieces of memory based on the shapes and data types of the tensors. To find suitable locations in the balloon, the algorithm compares the required sizes with available segments of free memory, starting from the beginning (i.e., first fit). Once it has found suitable pieces of memory, it marks them as allocated in the balloon and stops searching. During all steps, the algorithm keeps track of the maximum size of the memory balloon.  

The emitted code implements an API with two functions: The first function allows to set up the converted DNN and the second function executes an inference given an input sample. The latter is implemented based on the list of topologically sorted ONNX operator nodes stored in the previously generated intermediate representation. A function call is issued for each operator. These functions are implemented by the \emph{runtime library}. To give context to these functions and to pass intermediate results between them, we provide references to constant tensor descriptor structures generated as part of the intermediate representation. All static data associated with weight tensor descriptors is stored in a byte array in the intermediate representation. Therefore, a constant C array (i.e., flash memory) containing all the data is output accordingly. The amount of random access memory required for the intermediate activation tensors is based on the minimal memory balloon previously estimated by our \emph{memory planning algorithm}. Therefore, our tool issues another correspondingly sized zero initialized non-constant C array (i.e., heap memory).

\subsection{Runtime Library}
\label{sec:runtime library}

To perform the operations described by the input ONNX model, the code issued by our conversion tool relies on additional DNN operator functionality that we implement through a runtime library. Currently, this includes operators such as convolutions, linear transformations, batch normalization, pooling operations, and activation functions. All our implementations follow the ONNX operator specification\footnote{\url{https://github.com/onnx/onnx/blob/master/docs/Operators.md}}. Where necessary, we also implement quantized versions of these operators.

Depending on the target platform, there are several ways to optimize the execution of DNNs. During profile testing, we found that most of the execution time is spent computing convolutions or matrix-vector products. Therefore, an optimal implementation of these operator types leads to significant improvements in both resource consumption and execution time. Less crucial, but still significant, is that some operations can be removed from a DNN's computational graph through graph optimization, which we apply after DNN training and compression. Notable optimization techniques include \emph{batch normalization folding}~\cite{jacob_quantization_2017} and merging ReLU activation functions into preceding quantized linear or convolutional operations~\cite{jacob_quantization_2017}.

For our experiments, we focus on Cortex-M0+ and Cortex-M4 processors. Therefore, our implementation relies heavily on the information provided by these processor architectures. An important algorithmic optimization we apply is the unrolling of 2D convolutions into more CPU- and memory-friendly matrix-vector products (\emph{im2col mapping}). Thus, during mapping, we rearrange both the input tensors and the parameter tensors of the convolutions. This is a common approach in digital signal processing.\footnote{e.g., Matlab\textregistered{} provides \texttt{convmtx()} and \texttt{convmtx2()} for this task.}. In addition, this mapping also allows our conversion tool to apply CRS to convolutional slices.

Since DNNs also use matrix-vector products in linear transformations, a nice side effect is that by using the \textit{im2col} mapping, complete inference passes can be described by matrix-vector products and nonlinearities alone. In addition, ARM provides optimized open source implementations for matrix products in its \emph{CMSIS} library\footnote{\url{https://github.com/ARM-software/CMSIS_5}}. Using these is especially beneficial on architectures such as the Cortex-M4, as it allows the use of SIMD instructions provided by ARM's \emph{Digital Signal Processing (DSP)} extension, see section.~\ref{sec:eval}.
\section{Evaluation}
\label{sec:eval}

\begin{table}[ht!]
    \footnotesize
	\caption{Parameters of the DNN architectures used in our experiments.}
	\label{tab:architectures}
	\centering
	\begin{tabularx}{\linewidth}{ c Y Y }
		\toprule
		\textbf{AlexNet} (44.7M) & \textbf{ResNet} (9.4M) & \textbf{LeNet} (1.2M) \\
		\midrule
		\multicolumn{2}{c}{\textit{Input:} [3, 32, 32]} & \textit{Input:} [1, 28, 28]\\
		\midrule
		{[3, 64, 2, 2]} & {[64, 64, 3, 1]} & {[1, 32, 3, 1]} \\ 
		{[64, 192, 3, 1]} & {[64, 128, 3, 1]} & {[32, 64, 3, 1]} \\
		{[192, 384, 3, 1]} & {[128, 256, 3, 1]} & \\
		{[384, 256, 3, 1]} & {[256, 512, 3, 1]} & \\
		\midrule
		\multicolumn{3}{c}{MaxPool: [2]}\\
		\midrule
		{[6400, 4096]} & {[25088,10]} & {[9216, 128]} \\
		{[4096, 4096]} & & {[128, 10]} \\ 
		{[4096, 10]} & & \\ 
		\midrule
		\multicolumn{3}{c}{\textit{Output:} [10]}\\
		\midrule
		\multicolumn{3}{c}{SoftMax}\\
		\bottomrule
        \multicolumn{3}{p{\linewidth}}{\footnotesize Tuples describe linear layers in the form of [num. inputs, num. outputs]. Every layer uses ReLU as their non-linearity except the last ones which are followed by SoftMax; AlexNet/LeNet: quadruples describe 2D-convolutions by [channels, filters, kernel size, stride]; ResNet: quadruples describe residual blocks each with two 2D-convolutions in the form of [block in. channels, block out. filters, conv. kernel size, conv. stride]; AlexNet/ResNet: every 2D-convolution is followed by batch normalization.}
	\end{tabularx}
    \vspace{-4mm}%
\end{table}

\begin{table}[ht!]
    \footnotesize
    \caption{Microcontrollers considered in our evaluation.}
	\label{tab:target_systems}
	\centering
	\begin{tabularx}{\linewidth}{ l Y Y Y }
		\toprule
		{} & \textbf{Raspberry Pi Pico} & \textbf{Arduino Nano 33 BLE Sense} & \textbf{Raspberry Pi 4B} \\
		\midrule
		Processor & RP2040, Cortex-M0+ (Armv6-M) & nrf52840, Cortex-M4 (Armv7-M) & BCM2711 SoC, Cortex-A72 (ARMv8-A) \\
		Clock & 133 MHz & 64 MHz & 1.5 GHz \\
		Flash & 2 MB & 1 MB & 16 GB (SD-Card) \\
		RAM & 256 KB (SRAM) & 256 KB (SRAM) & 8 GB (SDRAM) \\
		SIMD & No & Yes (ARM DSP) & Yes (ARM Neon) \\
		\bottomrule
	\end{tabularx}
	\vspace{-4mm}%
\end{table}

To evaluate our pipeline, we selected three popular DNN architectures: (1) a convolutional network similar to the one proposed by Krizhevsky et al.~\cite{krizhevsky_imagenet_2017} to classify CIFAR-10 images (AlexNet), (2) a residual network~\cite{he_deep_2016} (ResNet), and (3) a smaller network architecture originally proposed by LeCun et al.~\cite{lecun_backpropagation_1989} (LeNet) to classify the MNIST handwritten digit database. See Table~\ref{tab:architectures} for more details.

We trained AlexNet and ResNet on the CIFAR10~\cite{krizhevsky_learning_2009} dataset for 100 epochs with mini-batches of size 80, and LeNet on the MNIST handwritten digit datasets~\cite{lecun_mnist_2010} for 20 epochs with mini-batches of size 48 (as training converges much faster on MNIST). For all models, we used stochastic gradient descent (SGD) with an impulse of 0.9 and a learning rate of $1e-3$.
We achieved a maximum accuracy of 86.51\% for AlexNet, 85.97\% for ResNet, and 98.46\% for LeNet, which serve as baselines for our experiments.

To evaluate our deployment pipeline, we selected three target systems: (1) a Raspberry Pi Pico, (2) an Arduino Nano 33 BLE Sense, and (3) a Raspberry Pi 4B (which serves as a larger reference system), see Table~\ref{tab:target_systems}. Since our runtime library is mainly optimized for Cortex-M architectures, we use onnxruntime~\cite{developers_onnx_2021} for the Raspberry Pi 4B instead.

We compare the performance of different pruning techniques in Sec.~\ref{subsec:eval-pruning} and discuss their combination with quantization in Sec.~\ref{subsec:eval-quantization}. We analyze the memory consumption of our compressed DNNs in Sec.~\ref{subsec:eval-memory}. In Sec.~\ref{subsec:eval-deploy}, we discuss the execution time and power/energy consumption vs. prediction accuracy from executions of the compressed DNNs on the target platforms.

\subsection{Comparison of Pruning Techniques}
\label{subsec:eval-pruning}

\begin{figure*}[t!]
    \centering
    \subfigure[AlexNet on CIFAR-10.]{
        \includegraphics[width=.28\textwidth]{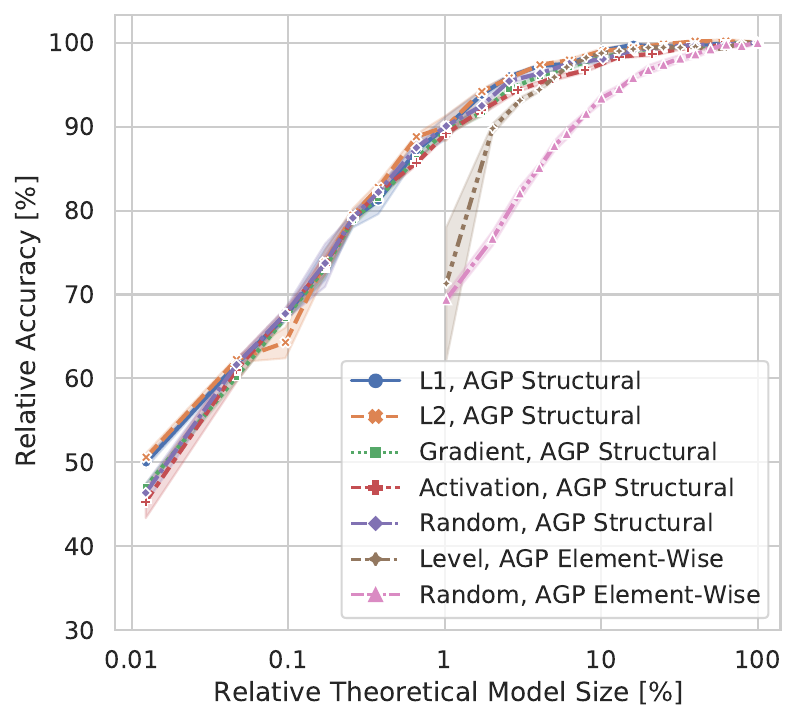}%
        \label{fig:pruning_alexnet}%
    }
    \subfigure[ResNet on CIFAR-10.]{
        \includegraphics[width=.28\textwidth]
            {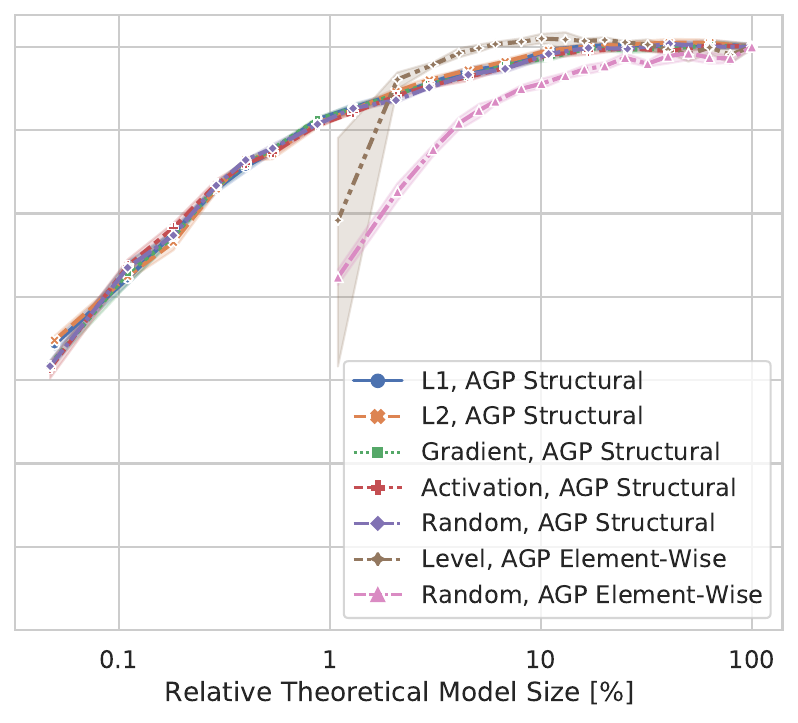}%
        \label{fig:pruning_resnet}%
    }
    \subfigure[LeNet on MNIST.]{
        \includegraphics[width=.28\textwidth]
            {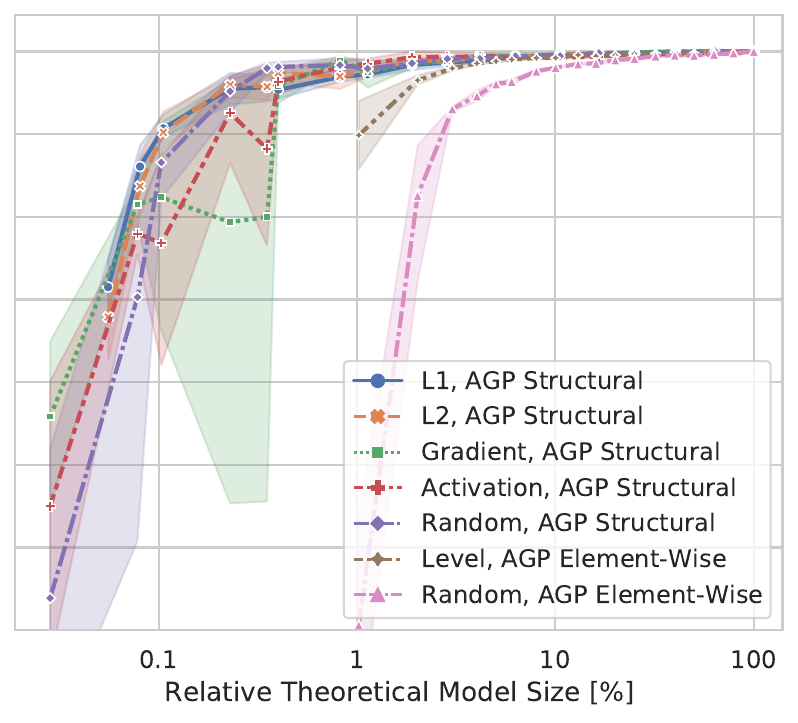}%
        \label{fig:pruning_lenet}%
    }
    \vspace{-1mm}%
    \caption{Element-wise and structural pruning applied to the DNN architectures. The curves describe the relation between theoretical model size and accuracy of compressed models relative to their uncompressed baselines when using different pruning techniques in combination with an iterative AGP pruning schedule. We define theoretical model size to be the number of weights a model features excluding all weights that have been set to zero by pruning.}
    \label{fig:pruning_results}
    \vspace{-4mm}%
\end{figure*}

First, we present the results of pruning experiments conducted for each of our three test DNN architectures. We repeated model training from scratch and increased the pruning target rates, starting with 0\% as the unpruned baseline and ending with 99\% (i.e., a relative theoretical model size of 1\%) as the most aggressively pruned configuration. For each of the configurations, we repeated the experiment five times and report their means and standard deviations.

Fig.~\ref{fig:pruning_results} shows the predictive accuracy of the models (relative to the un-pruned baseline) over percentages of theoretically remaining parameters on the validation dataset. 
Using an iterative AGP schedule, we tested different pruning heuristics for structural and element-wise pruning on all architectures, see Figs.~\ref{fig:pruning_alexnet} to~\ref{fig:pruning_lenet}. We report four different heuristics for structural pruning and one for element-wise pruning, along with a random selection approach as a baseline for the more sophisticated heuristics. For structural pruning, we use both the $\ell_1$ and $\ell_2$ norms of the parameter structures ('L1' and 'L2') as well as their gradient size ('Gradient') and the average percentage of zeros in their corresponding activation ('Activation') as heuristics. For element-wise pruning, we use a magnitude level to decide which elements to prune ('Level').

While we see a significant improvement of the level-based heuristic over the corresponding random selection for element-wise pruning on all our three DNN architectures, we cannot observe a similar behavior for structural pruning. Instead, none of the more complex structural pruning heuristics managed to significantly improve over the random selection approach. There is also little variation in the results of the heuristics. We believe that the main reason for this is the iterative retraining between pruning steps: while introducing pruning during DNN training can degrade the predictive quality of a model, it was very often recovered in a short number of epochs during retraining. This is consistent with results reported in previous work~\cite{han_learning_2015}.

In all our experiments, we used the same target compression rates for both element-wise and structural pruning. However, we see that the structural pruning experiments result in models that exceed their chosen target compression rate. In some cases, this reduces the number of parameters to almost 99.9\%. For element-wise pruning, we do not see such an effect. The reason for this is related to the removal of structures from DNN models during structural pruning: due to the existence of data dependencies between layers, removing structures from their parameter tensors also affects the shapes of tensors in surrounding layers. In element-wise pruning, parameters are not completely removed from the DNN, but only set to zero. Thus, all data dependencies remain in the network, tensor shapes do not change, and the pruning target rate is more accurately achieved.

\subsection{Combining Pruning and Quantization}
\label{subsec:eval-quantization}

\begin{figure}[t!]
    \centering
    \subfigure[L1-norm structural pruning]{\includegraphics[width=.49\linewidth]{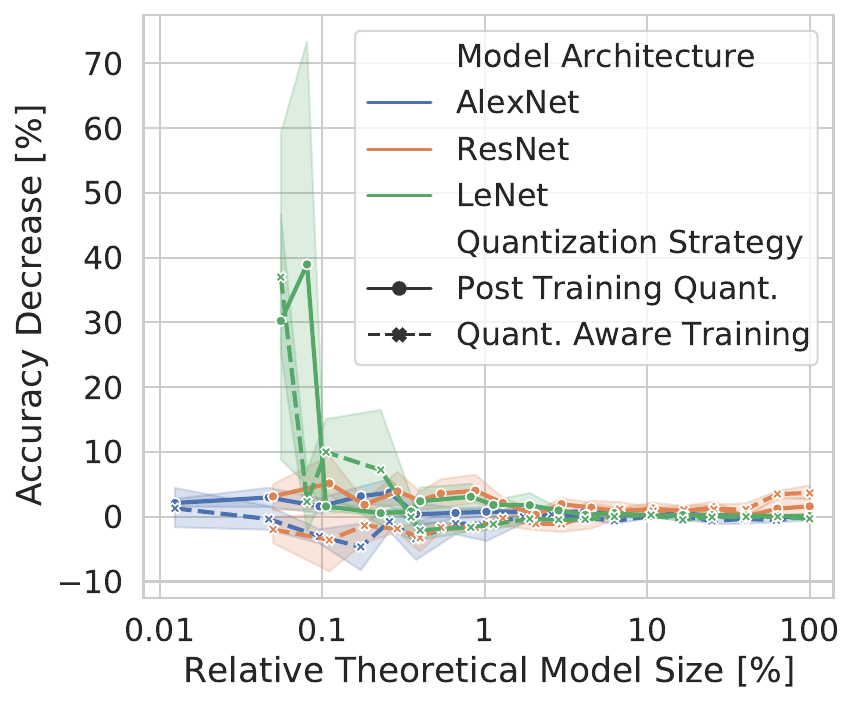}%
        \label{fig:quantization_results_structure}}
    \subfigure[Level element-wise pruning]{\includegraphics[width=.49\linewidth]{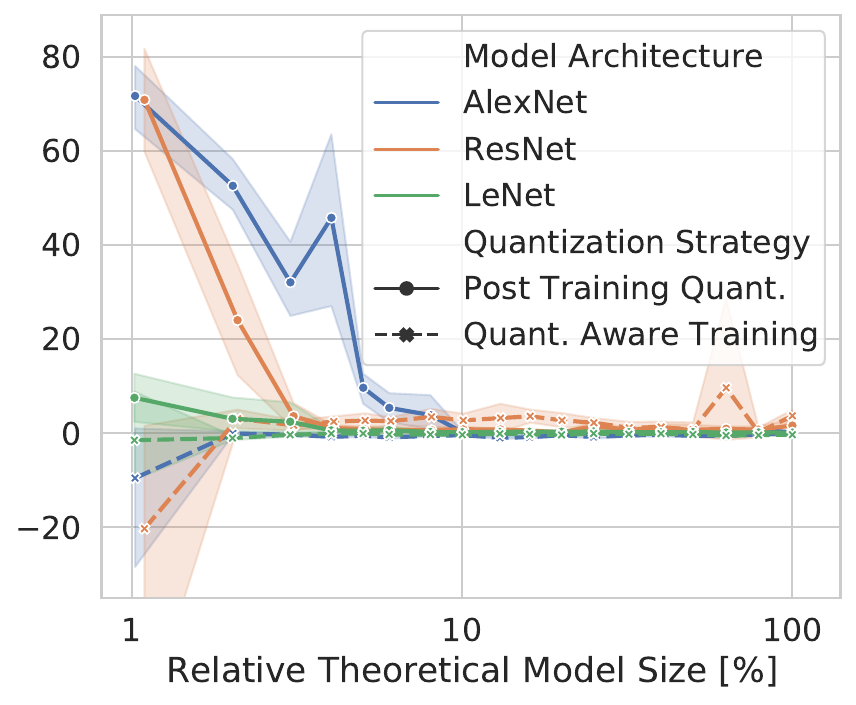}%
        \label{fig:quantization_results_element}}
    \vspace{-2mm}%
    \caption{Decrease in accuracy on the evaluation dataset observed for our three DNN architectures when combining pruning and quantization techniques.}
    \vspace{-4mm}%
    \label{fig:quantization_results}
\end{figure}

We present experimental results for weight quantization combined with pruning for our three test DNN architectures. Fig.~\ref{fig:quantization_results_structure} shows the results for quantization combined with structural pruning and Fig.~\ref{fig:quantization_results_element} shows the results for element-wise pruning. The different colors refer to the models we trained and the quantization strategies are distinguished by the line and marker style. We want to give an understanding of how much the additional quantization error alone affects the prediction quality of quantized models. Thus, as before, the $x$ axes show the relative theoretical model size reduction, while this time the $y$ axes show the accuracy reduction of each pruned and quantized model relative to its pruned but unquantized version.\footnote{We would like to point out that quantization does not change the number of parameters in a model, only their resolution, so applying quantization does not affect the relative theoretical model size, although the model does become effectively smaller. We present our results on the actual size reduction achievable with quantization in Sec.~\ref{subsec:eval-memory}.}. This allows us to focus on the additional error introduced by quantization alone.

Looking at the results of structural pruning in combination with both quantization as a post-process after training (PPQ) and quantization aware training (QAT), we see that these techniques work well with pruning for all three of our architectures. In Fig.~\ref{fig:quantization_results_structure} we see that the drop in accuracy is consistently $<5\%$ even when quantization is used in combination with aggressive pruning regimes. The only outliers we observed were part of our experiments on the LeNet architecture. Here, for the two most aggressive pruning configurations, the accuracy degradation between the unquantized and quantized models was about 40\% for both quantization strategies tested. In addition, we observed an increase in the standard deviation with decreasing accuracy, which we believe is related to an increase in the variance of the trained weight values that we observed for LeNet at higher compression rates. The higher the variance of the values in a weight tensor, the worse the quantization can represent those values in integer space.

We also tested element-wise pruning in combination with PPQ and QAT, see Fig.~\ref{fig:quantization_results_element}. Unlike structural pruning, where PPQ performed consistently well even in combination with aggressive pruning configurations, we observed a loss of accuracy of over 70\% for element-wise pruning compared to the unquantized versions. In particular, we observe that PPQ noticeably failed for models compressed by element-wise pruning to 10\% or less of their original parameter count. This is despite the fact that the technique performed well for pruning configurations targeting compression rates above 10\%. In contrast, QAT performed significantly better than PPQ even when used with aggressive pruning configurations. The technique was able to keep the loss of accuracy very close to 0\% in all the experiments conducted. Therefore, we conclude that PPQ seems to perform better when used in combination with structural pruning than when used with element pruning. QAT, on the other hand, performed consistently well, both in combination with structural and element-wise pruning.

\begin{figure*}[ht!]
    \centering
    \subfigure[Flash consumption of the compressed models (left: AlexNet/CIFAR-10; middle: ResNet/CIFAR-10; right: LeNet/MNIST).]{
        \includegraphics[width=.28\textwidth]{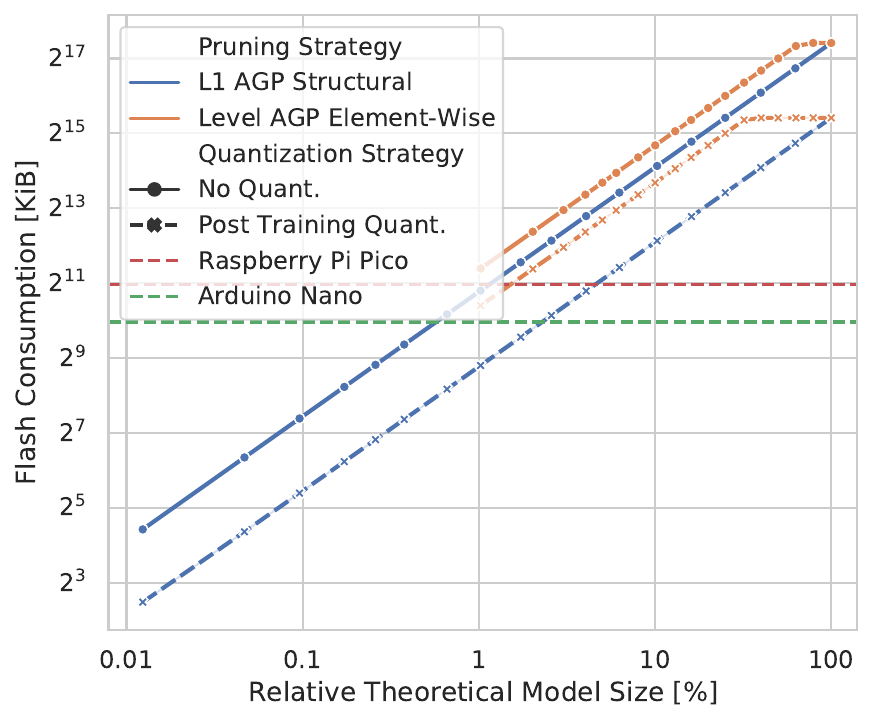}%
        \includegraphics[width=.28\textwidth]{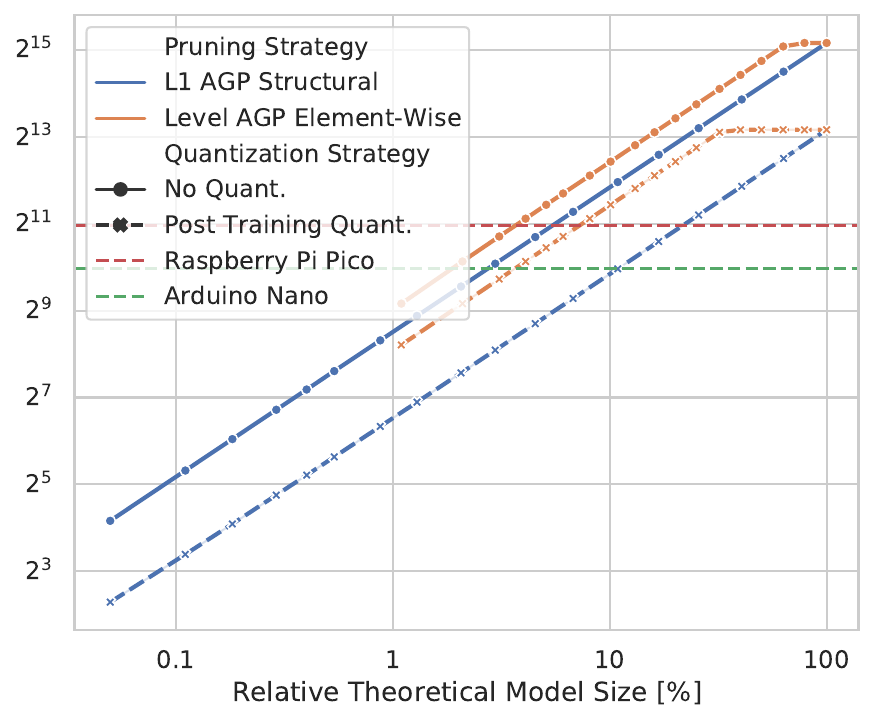}%
        \includegraphics[width=.28\textwidth]{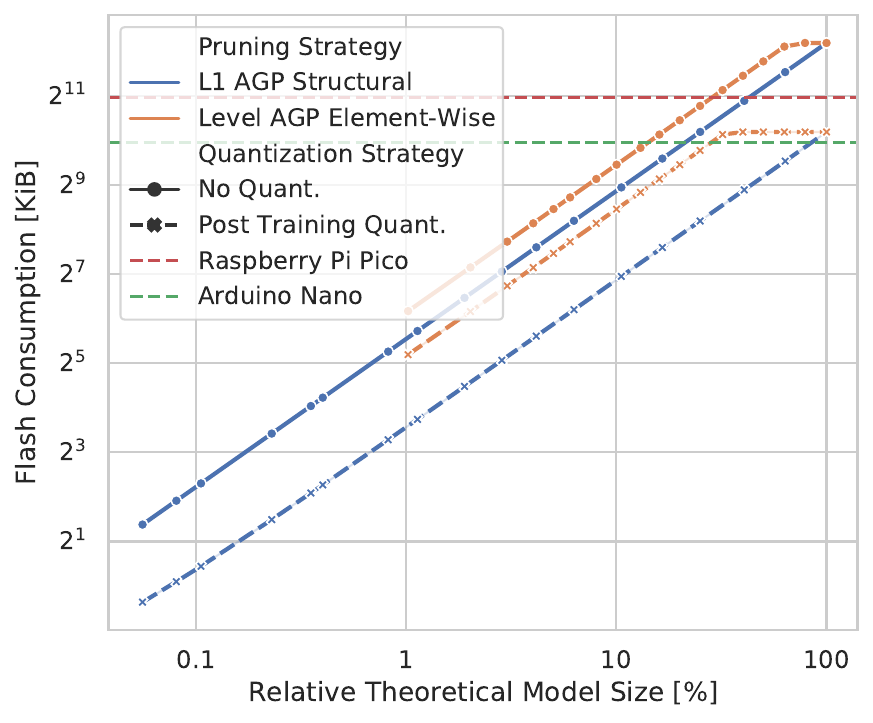}%
        \label{fig:size_results:flash}}
    \vspace{-2mm}%
    
    \subfigure[SRAM consumption of the compressed models (left: AlexNet/CIFAR-10; middle: ResNet/CIFAR-10; right: LeNet/MNIST).]{
        \includegraphics[width=.28\textwidth]{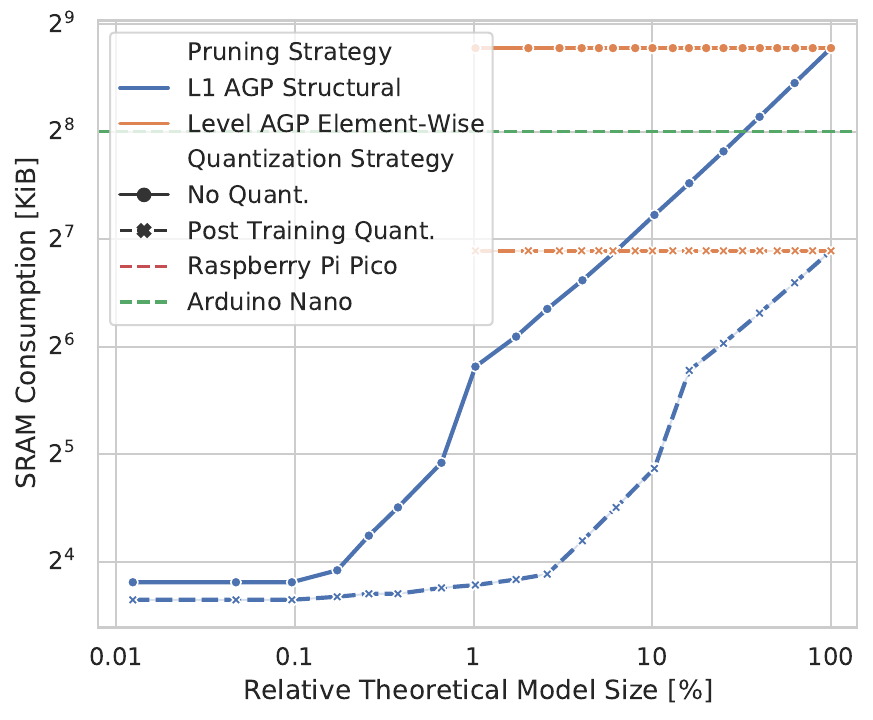}%
        \includegraphics[width=.28\textwidth]{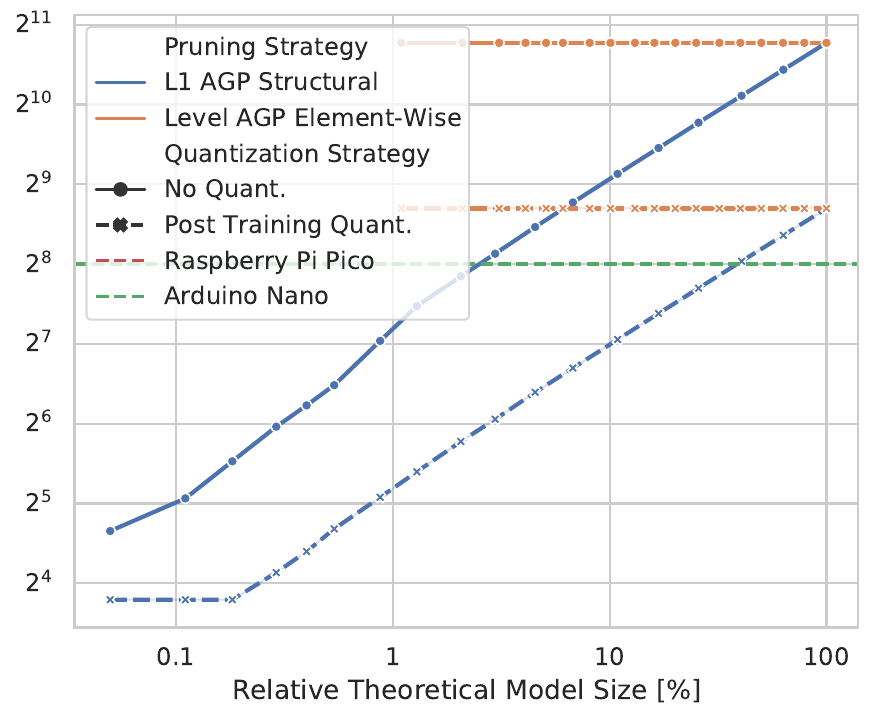}%
        \includegraphics[width=.28\textwidth]{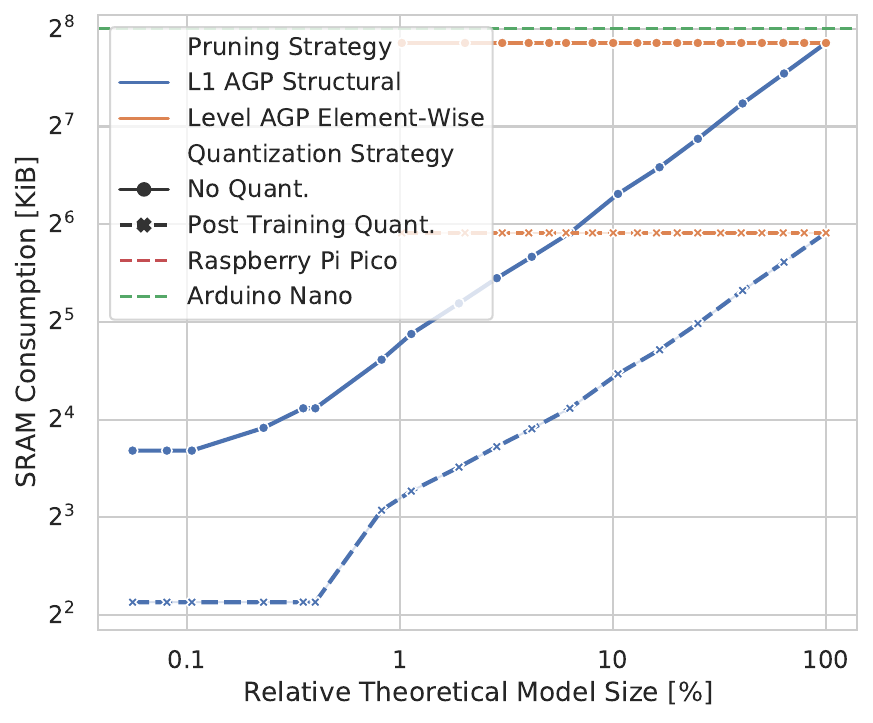}%
        \label{fig:size_results:sram}}
    
    \vspace{-1mm}
    \caption{Memory required for the deployment of our three DNN architectures (AlexNet, ResNet, LeNet) when applying the proposed runtime environment. (a) shows the required amounts of read-only "Flash" memory to store the DNNs' trained weights, while (b) shows the required amounts of random access memory "SRAM" to store the dynamic activation tensors. The dashed horizontal red and green lines mark the memory limits of our two considered embedded target platforms, the Raspberry Pi Pico and the Arduino Nano 33 BLE Sense.}
    \label{fig:size_results}
    \vspace{-4mm}%
\end{figure*}

\subsection{From Size Reduction to Memory Consumption}
\label{subsec:eval-memory}
    
As described in Sec.~\ref{sec:arch_depl}, the execution of DNNs requires both static read-only memory and dynamic random access memory. We now discuss the memory footprint of the models when they are deployed. Fig.~\ref{fig:size_results:flash} shows the relationship between relative model size (on the $x$-axes) and ROM/flash consumption in kibibytes (on the $y$-axes), and Fig.~\ref{fig:size_results:sram} shows the same relationship for SRAM. The curves for the pruned models are plotted with solid lines, while the curves for the models with additional quantization are plotted with dashed lines. To also highlight the importance of model compression, we have added the flash and SRAM limits of our two microcontroller target platforms (red and green dashed horizontal lines). On both of these platforms, flash and SRAM availability are two of the main bottlenecks to deployment.

In Fig.~\ref{fig:size_results:flash} we see a linear correlation between flash consumption and relative theoretical model size for structural pruning (note that both axes are on a logarithmic scale). The reason for this is that pruned structures can be completely removed from the model, which immediately reduces its memory consumption. For element-wise pruning, there is no such direct correlation. Instead, looking at the orange curves in Fig.~\ref{fig:size_results:flash}, we observe plateaus for higher theoretical model sizes. This is because pruning produces only sparse weight matrices. Our runtime environment uses compressed row storage (CRS) as a special decoding technique to store sparse weight matrices in a space-efficient way. However, a characteristic of the technique is that decoding a tensor only leads to memory savings after a certain percentage of sparsity has been reached. For any amount of sparsity below this threshold, it is better to just use the default storage layout where all elements are stored sequentially.

Another observation we made is that for models that are compressed using quantization and element-wise pruning, the threshold at which CRS becomes feasible is much higher than for models that are pruned alone. Again, this is related to the properties of CRS decoding. Instead of storing all values, only values unequal to zero (or a zero point) are stored. To preserve their position in the original undecoded matrix, the row and column indices of the values must also be stored. For larger matrices, such as those found in DNNs, these indices typically require 16- or 32-bit integers to store correctly. Therefore, the memory savings achieved by CRS can be considered as a trade-off between storing some elements and their indices and storing all elements without indices. When introducing sparsity into matrices with 32-bit floating-point values, this quickly becomes a good tradeoff. However, since quantized values require only 8 bits, while the index values introduced by CRS are usually still at least 16 bits long, the amount of sparsity that must be introduced before memory can be saved is greater.

Fig.~\ref{fig:size_results:sram} shows the relationship between relative theoretical model size and required SRAM. Similar to flash consumption, a correlation between model size and SRAM consumption can be observed for structure pruning. When structures are removed from parameter tensors during pruning, data dependencies between layers are also removed. Therefore, the shapes of the dynamic intermediate activation tensors stored in RAM and shared between subsequent layers are also reduced. The reason the relationship is not perfectly linear is that our runtime library tries to reuse heap memory for multiple activation tensors.  
We cannot observe a reduction in SRAM consumption for element-wise pruning. This is because we do not remove any elements during element-wise pruning. 

\begin{table*}[!t]
    \footnotesize
	\caption{Average ($n=8$) execution time, power and energy consumption observed per inference for compressed version of LeNet targeting two relative accuracy levels. The baseline model achieved an accuracy of $\approx$98\% for both float/uint8.}
	\label{tab:runtime_metrics_lenet}
	\centering
	\begin{tabular}{ l l l l S[table-format=4.2(4)] S[table-format=4.2(5)] S[table-format=3.2(3)] }
		\toprule
		rel. accuracy &compression & type & system & \textbf{execution time (ms)} & \textbf{power (mW)} & \textbf{energy (mWs)} \\
		\midrule
		\textbf{$>$99.00\%} & 99.18\% & structure, float & arduino & 44.15 \pm 0.03 & 141.14 \pm 0.07 & 6.23 \pm 0.01 \\
		{} & {} & {} & pico & 184.28 \pm 0.17 & 81.42 \pm 0.42 & 15.00 \pm 0.07 \\
		{} & {} & {} & pi4b & 1.81 \pm 0.55 & 3754.70 \pm 284.92 & 6.65 \pm 1.35 \\
		\hdashline
		{} & {} & structure, uint8 & arduino & 18.84 \pm 0.04 & 101.06 \pm 1.80 & 1.90 \pm 0.03 \\
		{} & {} & {} & pico & 50.06 \pm 0.03 & 79.37 \pm 1.03 & 3.97 \pm 0.05 \\
		{} & {} & {} & pi4b & 2.16 \pm 0.60 & 3624.90 \pm 272.88 & 7.68 \pm 1.53 \\
		\midrule
		{} & 95.00\% & element, float & arduino & {---} & {---} & {---} \\
		{} & {} & {} & pico & 1004.77 \pm 0.07 & 81.44 \pm 0.10 & 81.82 \pm 0.11 \\
		{} & {} & {} & pi4b & 5.66 \pm 1.39 & 3772.18 \pm 263.01 & 21.00 \pm 3.34 \\
		\hdashline
		{} & {} & element, uint8 & arduino & 360.95 \pm 0.09 & 142.28 \pm 0.10 & 51.36 \pm 0.04 \\
		{} & {} & {} & pico & 392.23 \pm 0.07 & 84.82 \pm 0.37 & 33.27 \pm 0.15 \\
		{} & {} & {} & pi4b & 5.10 \pm 1.27 & 3673.70 \pm 332.91 & 18.31 \pm 3.04 \\
		\midrule
		\textbf{$>$97.00\%} & 99.7\% & structure, float & arduino & 22.48 \pm 0.01 & 105.32 \pm 0.27 & 2.37 \pm 0.01 \\
		{} & {} & {} & pico & 73.26 \pm 0.33 & 81.72 \pm 0.48 & 5.99 \pm 0.03 \\
		{} & {} & {} & pi4b & 1.91 \pm 0.56 & 3554.33 \pm 232.34 & 6.67 \pm 1.48 \\
		\hdashline
		{} & {} & structure, uint8 & arduino & 18.79 \pm 0.03 & 82.54 \pm 0.80 & 1.55 \pm 0.02 \\
		{} & {} & {} & pico & 24.97 \pm 0.44 & 79.92 \pm 0.88 & 2.00 \pm 0.04 \\
		{} & {} & {} & pi4b & 2.01 \pm 0.53 & 3581.79 \pm 240.22 & 7.11 \pm 1.41 \\
		\midrule
		{} & 97.00\% & element, float & arduino & {---} & {---} & {---} \\
		{} & {} & {} & pico & 658.70 \pm 0.10 & 82.44 \pm 0.68 & 54.30 \pm 0.45 \\
		{} & {} & {} & pi4b & 5.35 \pm 1.23 & 3816.72 \pm 246.08 & 20.12 \pm 2.79 \\
		\hdashline
		{} & {} & element, uint8 & arduino & 303.15 \pm 0.09 & 141.58 \pm 0.05 & 42.92 \pm 0.01 \\
		{} & {} & {} & pico & 392.27 \pm 0.10 & 84.67 \pm 0.45 & 33.22 \pm 0.18 \\
		{} & {} & {} & pi4b & 4.64 \pm 1.39 & 3787.84 \pm 395.19 & 17.05 \pm 3.16 \\
		\bottomrule
	\end{tabular}
	\vspace{-4mm}%
\end{table*}

\subsection{Deployment Results}
\label{eval:deploy}
\label{subsec:eval-deploy}

As a final step in evaluating our pipeline, we deployed several of the pruned and quantized models from our previous experiments on our target systems and monitored key runtime metrics. In particular, we focused on execution time per inference, performance, and energy consumption.
To measure these metrics on our test systems, we used a \textit{Agilent N6705A DC Power Analyzer} to provide them with a regulated power supply. The power analyzer also allowed us to measure the current and power consumed by the systems. To measure the execution time required to compute the power consumption of our DNN models, we used a GPIO signal. We toggled the signal at the start and end of each inference and monitored it with an oscilloscope. We present the results of our measurements for LeNet in Table~\ref{tab:runtime_metrics_lenet} and the results for our other two DNN architectures in Table~\ref{tab:runtime_metrics}.

\begin{table*}[ht!]
    \footnotesize
	\caption{Average ($n=8$) execution time, power and energy consumption observed per inference for AlexNet and ResNet. To achieve different compression rates and thus different accuracy scores, we used L1-norm structural pruning. In addition, we used post training quantization for all non floating-point measurements.}
	\label{tab:runtime_metrics}
	\centering
	\subfigure[AlexNet, CIFAR10, baseline accuracy $\approx$86\% for both float and uint8]{%
			\begin{tabular}{ c l c l S[table-format=4.2(4)] S[table-format=4.2(5)] S[table-format=3.2(3)] }
				\toprule
				compression & type & rel. accuracy & system & \textbf{execution time (ms)} & \textbf{power (mW)} & \textbf{energy (mWs)} \\
				\midrule
				\textbf{95.9\%} & float & 97.79\% & arduino & {---} & {---} & {---} \\
				{} & {} & {} & pico & {---} & {---} & {---} \\
				{} & {} & {} & pi4b & 6.60 \pm 1.85 & 3795.81 \pm 334.67 & 24.43 \pm 4.10 \\
				\hdashline
				{} & uint8 & 97.56\% & arduino & {---} & {---} & {---} \\
				{} & {} & {} & pico & 1766.58 \pm 0.25 & 92.40 \pm 0.03 & 163.23 \pm 0.07 \\
				{} & {} & {} & pi4b & 3.96 \pm 1.12 & 3809.41 \pm 250.59 & 14.82 \pm 2.71 \\
				\midrule
				\textbf{98.2\%} & float & 94.88\% & arduino & {---} & {---} & {---} \\
				{} & {} & {} & pico & {---} & {---} & {---} \\
				{} & {} & {} & pi4b & 2.68 \pm 0.70 & 3687.37 \pm 221.25 & 9.71 \pm 1.76 \\
				\hdashline
				{} & uint8 & 94.65\% & arduino & 403.11 \pm 0.16 & 139.96 \pm 0.17 & 56.42 \pm 0.07 \\
				{} & {} & {} & pico & 669.97 \pm 0.19 & 86.41 \pm 0.02 & 57.89 \pm 0.03 \\
				{} & {} & {} & pi4b & 2.93 \pm 0.41 & 4187.75 \pm 89.73 & 12.28 \pm 2.05 \\
				\midrule
				\textbf{99.6\%} & float & 83.26\% & arduino & 252.40 \pm 0.40 & 107.83 \pm 0.42 & 27.22 \pm 0.10 \\
				{} & {} & {} & pico & 1435.42 \pm 0.34 & 81.55 \pm 0.07 & 117.06 \pm 0.09 \\
				{} & {} & {} & pi4b & 2.45 \pm 0.09 & 4182.59 \pm 66.35 & 10.25 \pm 0.35 \\
				\hdashline
				{} & uint8 & 82.33\% & arduino & 110.00 \pm 0.01 & 141.17 \pm 0.04 & 15.53 \pm 0.01 \\
				{} & {} & {} & pico & 186.57 \pm 0.12 & 86.86 \pm 0.21 & 16.21 \pm 0.04 \\
				{} & {} & {} & pi4b & 2.10 \pm 0.14 & 4134.21 \pm 52.03 & 8.68 \pm 0.60 \\
				\bottomrule
		\end{tabular}
		\label{tab:runtime_metrics_alexnet}}
	\hfill
	\subfigure[ResNet, CIFAR10, baseline accuracy $\approx$85\% for float and $\approx$84\% for uint8]{%
		\begin{tabular}{ c l c l S[table-format=5.2(4)] S[table-format=4.2(5)] S[table-format=4.2(5)] }
				\toprule
				compression & type & rel. accuracy & system & \textbf{execution time (ms)} & \textbf{power (mW)} & \textbf{energy (mWs)} \\
				\midrule
				\textbf{93.2\%} & float & 96.25\% & arduino & {---} & {---} & {---} \\
				{} & {} & {} & pico & {---} & {---} & {---} \\
				{} & {} & {} & pi4b & 29.41 \pm 2.62 & 4234.81 \pm 121.13 & 124.24 \pm 7.11 \\
				\hdashline
				{} & uint8 & 96.43\% & arduino & 11983.40 \pm 2.05 & 143.74 \pm 0.49 & 1722.48 \pm 5.69 \\
				{} & {} & {} & pico & 25997.20 \pm 4.00 & 91.94 \pm 0.01 & 2390.24 \pm 0.41 \\
				{} & {} & {} & pi4b & 21.32 \pm 3.32 & 4085.92 \pm 213.90 & 86.54 \pm 9.62 \\
				\midrule
				\textbf{98.7\%} & float & 92.50\% & arduino & 6105.20 \pm 1.60 & 105.76 \pm 1.85 & 645.69 \pm 11.16 \\
				{} & {} & {} & pico & 46266.50 \pm 16.82 & 85.78 \pm 0.02 & 3968.83 \pm 1.49 \\
				{} & {} & {} & pi4b & 9.41 \pm 4.27 & 4536.99 \pm 183.46 & 42.06 \pm 16.53 \\
				\hdashline
				{} & uint8 & 88.52\% & arduino & 2425.50 \pm 0.92 & 102.38 \pm 1.10 & 248.31 \pm 2.70 \\
				{} & {} & {} & pico & 4269.78 \pm 2.58 & 87.81 \pm 0.04 & 374.92 \pm 0.36 \\
				{} & {} & {} & pi4b & 7.74 \pm 2.78 & 3863.07 \pm 427.70 & 28.82 \pm 6.98 \\
				\midrule
				\textbf{99.6\%} & float & 83.48\% & arduino & 1754.75 \pm 0.31 & 106.57 \pm 2.64 & 187.00 \pm 4.63 \\
				{} & {} & {} & pico & 11140.25 \pm 1.85 & 83.13 \pm 0.24 & 926.12 \pm 2.61 \\
				{} & {} & {} & pi4b & 5.42 \pm 1.66 & 3560.66 \pm 421.12 & 18.70 \pm 4.01 \\
				\hdashline
				{} & uint8 & 78.28\% & arduino & 806.80 \pm 0.60 & 104.48 \pm 0.09 & 84.29 \pm 0.13 \\
				{} & {} & {} & pico & 1214.53 \pm 1.00 & 84.13 \pm 0.06 & 102.18 \pm 0.10 \\
				{} & {} & {} & pi4b & 3.45 \pm 0.22 & 3881.06 \pm 72.77 & 13.37 \pm 0.60 \\
				\bottomrule
		\end{tabular}
		\label{tab:runtime_metrics_resnet}}
		\vspace{-4mm}%
\end{table*}

For the LeNet architecture, we used the accuracy of the models on its evaluation dataset to compare the use of different pruning and quantization techniques. The logic behind this was that if models compressed using different approaches can achieve similar accuracy, then they can be considered direct alternatives and are therefore comparable. In our experimental setup, we defined two relative accuracy bounds for which we selected the smallest compressed models from our previous tests that met them: $>$99\% and $>$97\%. In the second column of Table~\ref{tab:runtime_metrics_lenet} we can see that LeNet can be compressed to well over 5\% of its original parameter count and still pass both accuracy bounds. We tested all selected models on our three target systems, including not only the pruned models but also their quantized counterparts, see the third and fourth columns. Note that using element pruning, we were not able to deploy all selected models on the Arduino.

For all deployed LeNet models, we monitored execution time, performance, and energy consumption over a range of 8 different conclusions. The resulting averages are shown in the remaining columns of Table~\ref{tab:runtime_metrics_lenet}. First, we see that the execution time per inference on the Pi 4B is significantly lower than on the Arduino or the Pi Pico. This is to be expected since the Pi 4B runs between 1.0 and 1.5 GHz, while both the Arduino and the Pico run in the lower MHz range. When comparing the Raspberry Pi Pico and the Arduino Nano, we observed a higher execution time on the Pico than on the Arduino (consistently). This is despite the fact that the Arduino runs at about half the clock speed of the Pico. On both systems, the quantized models always outperformed their floating-point counterparts, although much more so on the Pico. This can be explained by features present on both systems. First, the Arduino's Cortex-M4 processor implements a real floating-point unit, while the Pi Pico's Cortex-M0+ processor has to simulate floating-point arithmetic. Second, the Arduino's Cortex-M4 processor supports ARM's DSP extension, which gives it access to a subset of SIMD instructions to speed up its integer operations. The Pi Pico does not implement the DSP extension.

Considering the power measured during the inference for all the models used, we see that all our tested systems consume on average a constant amount of power, while we see a more significant variation in the different samples taken for each model on the Pi 4B (note that the power consumption is much higher than on the other two systems). On the Arduino, the measured power consumption was between 100 and 150 mW on average, while on the Pi Pico it was around 80 mW. In contrast, the Pi 4B generally consumed about 4W. However, in addition to power, execution time is the second factor in calculating a system's energy consumption. Looking at the results, we see that in some cases the Pi 4B has the best energy consumption per conclusion. It is often followed by the Arduino and then the Pi Pico. This is the opposite of power consumption and shows that faster inference can compensate for high power consumption.

Besides LeNet, we also used our AlexNet and ResNet architectures, see Table~\ref{tab:runtime_metrics}. For both architectures, we evaluated models that were compressed using structural pruning. This is different from our LeNet experiments, where we tested both structural and element-wise pruning. The reason we did not do this for AlexNet and ResNet is that memory limitations on our Arduino and Pi Pico target systems made it impossible for us to use element-wise pruned models. For structural pruning, the situation is different, and we were able to deploy feasible models. However, we were still forced to select models trained with aggressive compression rates that removed well over 90\% of the original parameters. As a result, we had to sacrifice accuracy, see the third column of Tables~\ref{tab:runtime_metrics_alexnet} and \ref{tab:runtime_metrics_resnet}.

To measure execution time, performance, and energy consumption, we used the same approach as before. We again monitored all three metrics over 8 different inferences and calculated their averages and respective standard deviations. Looking at the results for AlexNet and ResNet, we see the same patterns we discussed for LeNet in Table~\ref{tab:runtime_metrics_lenet}. However, we evaluated the two architectures as a way to explore the upper limits of feasible DNN deployment. AlexNet has a very large number of trainable parameters, while ResNet has a large number of large and computationally expensive convolutions. This has different implications for the deployment of the two architectures. While for AlexNet we need to apply high compression rates to shrink the model size enough to fit into the memory of our target microcontrollers (see Sec.~\ref{subsec:eval-memory}), for ResNet execution time is the primary bottleneck.
We conclude that it is not only the number of parameters of a model, but also its topology that determines whether it can be used on a target system from a performance point of view.

\begin{table*}[ht!]
    \footnotesize
    \centering
	\caption{Comparison between metrics measured when deploying AlexNet using tfl-micro. The percentage columns denote how much a metric increased on average when using tfl-micro instead of our toolflow.}
	\begin{tabular}{ c l l S[table-format=4.2(3)] S[table-format=4.2] S[table-format=3.2(3)] S[table-format=4.2] S[table-format=3.2(3)] S[table-format=4.2] }
		\toprule
		compression & type & system & \textbf{latency (ms)} & {(\%)} & \textbf{power (mW)} & {(\%)} & \textbf{energy (mWs)} & {(\%)} \\
		\midrule
		\textbf{98.2} & float & arduino & {---} & {} & {---} & {} & {---} & {} \\
		{} & {} & pico & {---} & {} & {---} & {} & {---} & {} \\
		\hdashline
		{} & int8 & arduino & 355.55 \pm 0.07 & -11.80 & 118.25 \pm 0.55 & -15.51 & 42.04 \pm 0.19 & -25.49 \\
		{} & {} & pico & 693.57 \pm 0.18 & 3.52 & 90.07 \pm 0.02 &  4.24 & 62.47 \pm 0.01 & 7.91 \\
		\midrule
		\textbf{99.6} & float & arduino & 595.75 \pm 0.19 & 136.00 & 150.00 \pm 0.55 & 39.11 & 89.36 \pm 0.32 & 228.29 \\
		{} & {} & pico & 1842.05 \pm 0.32 & 28.33 & 83.48 \pm 0.01 & 2.37 & 153.78 \pm 0.04 & 31.37 \\
		\hdashline
		{} & int8 & arduino & 95.34 \pm 0.09 & -13.33 & 151.16 \pm 0.02 & 7.10 & 14.41 \pm 0.01 & -7.22 \\
		{} & {} & pico & 172.76 \pm 0.10 & -7.40 & 89.21 \pm 0.03 &  2.71 & 15.41 \pm 0.01 & -4.94 \\
		\bottomrule
	\end{tabular}
	\label{tab:tflmicro_results}
    \vspace{-4mm}
\end{table*}

As a final contribution, we compare our deployment process with another option available for deploying DNNs on limited hardware. One of the most prominent alternatives to our pipeline is tfl-micro \cite{MLSYS2021_tflmicro}. The library is integrated into the tensorflow framework and provides the necessary capabilities to run trained tensorflow or keras models on ARM Cortex-M based hardware. To compare our implementation with that of tfl-micro, we use the same AlexNet architecture we introduced earlier as an example. However, keras provides only limited options for pruning DNNs. Therefore, we decided to take our existing AlexNet models, which we pruned using structure pruning, and convert them to keras. For quantization, tensorflow implements its own version of PTSQ (with an int8 mapping instead of uint8). This allowed us to run the same models using both our pipeline and tfl-micro's pipeline, allowing a direct comparison with Fig.~\ref{tab:runtime_metrics_alexnet}.

In Fig.~\ref{tab:tflmicro_results} we present the results we measured for deploying AlexNet using tfl-micro. We also show the percentage increase in latency, energy, and power consumption for all results obtained with tfl-micro compared to the results obtained with our own deployment library. A positive percentage means that a higher value for a metric was measured when using tfl-micro instead of our solution, while a negative percentage means that a lower value was observed. Comparing tfl-micro with our results, we see that our deployment library performs significantly better than tfl-micro for floating-point models. This is especially visible on the Arduino Nano, where we observed a 136\% increase in runtime when using our toolflow instead of tfl-micro. For quantized models, tfl-micro slightly outperformed our implementation. Regarding power consumption, all results show that tfl-micro almost always consumed more power than our approach. One reason may be that tfl-micro interprets the model and allocates memory dynamically at runtime. 

\section{Discussion of Results}

Using our compression and deployment pipeline, we were able to automatically and feasibly deploy DNNs on microcontrollers. By compressing DNNs, we were able to achieve significant savings in memory, execution time, and runtime energy consumption without sacrificing model accuracy. For pruning, we achieved the best results and most savings with structural pruning. When comparing different pruning strategies, our experiments indicate that structural pruning offers better opportunities for saving memory and execution time than element-wise pruning. In addition, the execution of DNNs compressed using this technique did not require special support for sparse matrix formats, as is required for element pruning. Furthermore, we observed that the use of different state-of-the-art heuristics did not have much impact on the structural pruning. Choosing a reasonable pruning schedule and allowing for retraining proved to be more effective. In addition, using weight quantization together with structural pruning resulted in even more savings not only in memory consumption, but also in execution time. This is due to the fact that our target systems were able to process much more efficiently in integer than in floating-point space. Furthermore, we found that our different compression and deployment strategies had almost no impact on the power consumed by both the Pi Pico and the Arduino during inference. 
This means that the observed energy savings were mainly the result of execution times.

%
We conclude that a DNN model is optimally deployed on a microcontroller if it runs on a system where it fits into the available memory and consumes the least amount of power under load, while still being able to perform inference in a reasonable time frame. Furthermore, we argue that the execution time of a DNN must be seen in relation to the frequency with which input samples are generated by the connected sensors.
\section{Conclusion}
\label{sec:conclusion}

In this work, we have presented a configurable pipeline for compressing DNNs and deploying them on Cortex-M-based microcontrollers. To achieve compression, our pipeline uses network pruning and weight quantization techniques. Deployment is handled by a proposed runtime environment, which consists of a code generator for mapping trained networks to executable C code and a runtime library that provides optimized implementations of common DNN layers. We used the introduced pipeline to compare DNNs compressed with different pruning and quantization techniques. We also tested how compression affects runtime performance on several target systems. We were able to show that even larger DNN architectures such as AlexNet or ResNet can be feasibly deployed on microcontrollers with memory footprints as small as 1-2 MB Flash and 256 Kb SRAM, while still achieving good execution time and accuracy results.

\section*{Acknowledgement}
This work was partially supported by the Bavarian Ministry for Economic Affairs, Infrastructure, Transport and Technology through the Center for Analytics-Data-Applications (ADA- Center) within the framework of “BAYERN DIGITAL II” and by the Deutsche Forschungsgemeinschaft (DFG, German
Research Foundation) under grant 45098171.

\bibliographystyle{IEEEtran}
{\footnotesize
\bibliography{literature.bib}}

\end{document}